\documentclass[11pt]{article}
\usepackage[final]{acl}
\usepackage{amsmath} 
\usepackage{times}
\usepackage[T1]{fontenc}
\usepackage{multirow}
\usepackage{booktabs}
\usepackage{hyperref}
\usepackage{cleveref}
\usepackage{twemojis}
\usepackage{linguex}
\usepackage[dvipsnames]{xcolor}
\usepackage{amssymb}

\usepackage{times}
\usepackage{latexsym}
\usepackage{booktabs}
\usepackage{multicol}
\usepackage{multirow}
\usepackage{hyperref}
\usepackage{adjustbox}

\usepackage{tikz-dependency}
\definecolor{yellowgreen}{RGB}{154, 205, 50}

\usepackage{microtype}

\usepackage{inconsolata}

\usepackage{stackengine}
\newcommand{\pmark}{%
  \textcolor{yellowgreen}{%
    \stackon[-3pt]{$\checkmark$}{\rule{0.8em}{0.4pt}}%
  }%
}

\usepackage{graphicx}

\usepackage{pifont}
\newcommand{\cmark}{\textcolor{green!60!black}{\ding{51}}}
\newcommand{\xmark}{\textcolor{red}{\ding{55}}}

\newif\ifshowemoji
\showemojitrue   

\newcommand{\bemoj}[1]{\ifshowemoji\texttwemoji{#1}\fi}

\title{Apples to Apples? Towards Comparable Crosslingual \\ Language Model Evaluation}

\author{
    {\bf Xiulin Yang}~\bemoj{apple} \quad 
    {\bf Ethan Gotlieb Wilcox}~\bemoj{apple}  \quad
    {\bf Catherine Arnett}~\bemoj{orange} \quad \\ 
\bemoj{apple} Georgetown University \bemoj{orange} EleutherAI\\
 \small{\texttt{\{xy236,ethan.wilcox\}@georgetown.edu, catherine@eleuther.ai
 }}
}

\begin{document}
\maketitle
\begin{abstract}
Crosslingual evaluation of language models that enables fair comparisons remains a fundamental challenge in multilingual NLP. Existing studies adopt a variety of downstream tasks and intrinsic metrics with different theoretical justifications, yet there has been little empirical investigation into whether these approaches yield meaningful crosslingual conclusions. We systematically examine crosslingual evaluation approaches using controlled monolingual language models trained on parallel data with varying tokenizer vocabulary sizes and model sizes, and further validate our findings on multilingual LLMs. We further discuss challenges in achieving comparable downstream evaluation across languages. Our results show that several widely used normalized metrics introduce crosslinguistic biases rooted in tokenization, encoding, and orthographic differences. In contrast, sentence-level negative log-likelihood computed over semantically equivalent sequences provides more meaningful and consistent crosslingual comparisons.\footnote{Code:\url{https://github.com/xiulinyang/multilingual-eval.git}; models:\url{https://huggingface.co/parallelm}. To load the tokenizer, please use \texttt{PreTrainedTokenizerFast} rather than \texttt{AutoTokenizer}.}\looseness=-1

\end{abstract}
\section{Introduction}
\label{sec:intro}
Fairly evaluating language models across typologically diverse languages is a fundamental yet underappreciated challenge in multilingual NLP. Despite significant progress in extending language technology to an ever-growing number of languages for both monolingual \citep[e.g.,][]{aravinda2025sinllama,poro2_2025} and multilingual models \citep[e.g.,][]{workshop2022bloom}, substantial performance gaps across languages persist \citep{chang2023multilinguality,joshi-etal-2020-state,shani2026roots}. Yet accurately measuring these gaps itself is a challenge: different metrics can yield substantially different conclusions about crosslingual model performance \citep{shani2026roots}, raising the question of what a \emph{fair} comparison even means.

Evaluation approaches in multilingual NLP broadly fall into two categories. Downstream task evaluations assess a model across a range of linguistic competencies, but fair crosslingual comparison using such evaluations is far from straightforward: conclusions are highly task-dependent \citep{rust-etal-2021-good,limisiewicz-etal-2023-tokenization}, and language-specific benchmarks may not exist for many languages. Intrinsic metrics, by contrast, directly quantify how well a model assigns probability to held-out parallel texts, offering a more feasible and controlled signal for crosslingual comparison. Yet, as we discuss below, ``more controlled'' does not necessarily mean ``fair''.

\begin{figure}[t]
    \centering
    \includegraphics[width=\linewidth]{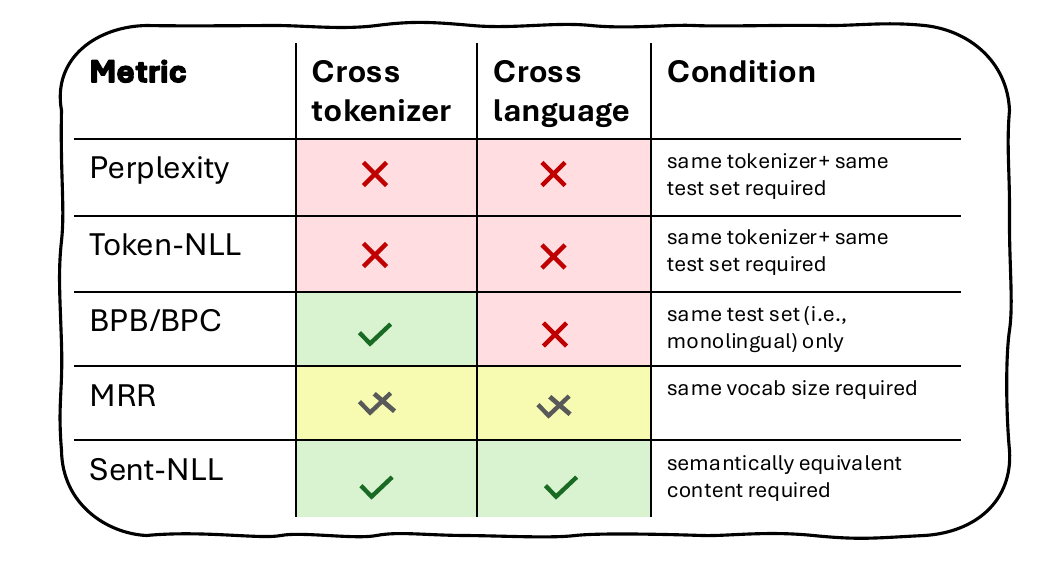}
    \caption{Suitability of intrinsic metrics for cross-tokenizer and cross-lingual comparison (\cmark~= suitable; \xmark~= unsuitable; \pmark~= conditionally suitable). Sentence NLL computed on semantically equivalent parallel content is the only metric suitable for both comparisons.}
    \label{fig:metric_suggesetion}
\end{figure}

Within intrinsic evaluations, there is no consensus on what probability-based metric to use. One line of work evaluates the probability assigned to sequences that express the same meaning across languages, typically using negative log-likelihood, surprisal, or related derived metrics \citep[e.g.,][]{mielke-etal-2019-kind, cotterell-etal-2018-languages, wan2022fairness}. Another common approach relies on unit-normalized variants such as perplexity (PPL; \citealp{shliazhko-etal-2024-mgpt,arnett-bergen-2025-language,thakur2025art}), bits-per-byte (BPB; \citealp{xue2022byt5}), or bits-per-character (BPC; \citealp{blevins-etal-2022-analyzing}), which normalize likelihood by tokens, bytes, or characters, respectively. Despite the intuitive justification for each of these metrics (e.g., BPB is tokenizer-agnostic; \citealp{gao2020pile}), empirical evidence examining these claims about which metric to use has remained limited. Recently, \citet{poelman2026form} argue that current evaluation metrics may not be universally comparable for two reasons: (i) different metrics can produce different rankings across languages, and (ii) even within a single metric, paraphrases expressing the same meaning can lead to inconsistent crosslingual rankings. These findings leave open the broader question of what constitutes a fair crosslingual comparison. \looseness=-1

In this paper, we argue that a fair metric is one satisfying three criteria: first, it should be \textit{not systematically driven by engineering choices}, specifically tokenization, encoding, and orthography. That is, language rankings should reflect language modeling quality rather than representation-level factors such as the number of characters, bytes, or tokens. Second, it should be \textit{robust to variation across semantically equivalent forms}: language rankings should remain stable across translation alternatives expressing the same content. Third, it should be \textit{model-general}, meaning that its behavior should generalize across monolingual and multilingual models of different scales.

To isolate metric behavior from confounds inherent in multilingual models, we first use 50  monolingual language models trained on parallel corpora with five vocabulary size settings \citep{yang2026} and conduct a systematic empirical comparison of six intrinsic metrics, then extend this comparison to larger monolingual and multilingual models. We find that five out of the six metrics violate at least one criterion above. By contrast, sentence-level NLL, which normalizes by meaning rather than by surface-level units such as morphemes or characters, satisfies all three: it is (i) substantially less sensitive to tokenization and encoding choices (\S\ref{exp1}), (ii) more stable across translation alternatives (\S\ref{exp2}), and (iii) applicable to both monolingual and multilingual settings across scales (\S\ref{exp3}). Our results challenge both common practice and recent proposals in multilingual evaluation literature. We show that both token-normalized metrics such as perplexity and character- and byte-level alternatives, which have been explicitly advocated as more robust \citep{shani2026roots}, introduce substantial and systematic biases that favor languages with longer sequences or more tokens. At the same time, we empirically validate the theoretically motivated assumption underlying prior work that uses sentence-level NLL or related transformed metrics \citep[e.g.,][]{mielke-etal-2019-kind, cotterell-etal-2018-languages, chang2023multilinguality}, providing evidence for claims that had previously lacked direct empirical support. We further discuss considerations for downstream evaluation, noting the challenges of achieving comparable assessment across languages and tasks.



\section{Background \& Related Work}
\label{related_work}

\subsection{Tokenization \& Representation Biases}
For crosslinguistic evaluation to be meaningful, it is necessary to disentangle intrinsic linguistic complexity from technical artifacts introduced by representation and preprocessing choices \citep{shani2026roots}. Prior work has identified several sources of such artifacts, particularly those arising from tokenization and orthographic encoding \citep[e.g.,][]{petrov2023language,ahia2024magnet}.

\paragraph{Tokenization}
Language models are trained to predict the next token in a sequence, from a vocabulary of discrete tokens. Consequently, how text is segmented into tokens directly affects how the model represents and learns linguistic structure. Semantically equivalent content across languages can be segmented into very different numbers of tokens, even when the same tokenization algorithm is used \citep{petrov2023language,arnett2025explaining}. This disparity is often quantified with \textit{Corpus Token Count} (CTC; \citealp{schmidt-etal-2024-tokenization}), which measures the total number of tokens required to encode a corpus under a given tokenizer.

Differences in token segmentation can influence both training dynamics and evaluation metrics. Even under identical tokenization settings, languages may be split into substantially different numbers of segments, which can affect model performance \citep{shani2026roots}. As a result, different languages may benefit from different tokenization strategies \citep[e.g.,][]{fujii-etal-2023-different,vemula-etal-2025-rethinking,reddy2025much,toraman2023impact}. For instance, \citet{toraman2023impact} show that morphologically-rich languages like Turkish require a larger vocab size to train and fine-tune encoder models.


\paragraph{Orthography and Encoding}
Representation disparities also arise from differences in orthographic encoding. The same semantic content written in different scripts may require different numbers of UTF-8 bytes, a phenomenon known as the \textit{byte premium} \citep{arnett-etal-2024-bit}. Similarly, different writing systems may encode equivalent content using substantially different numbers of characters, resulting in what has been termed the \textit{length premium} \citep{arnett-etal-2024-bit}. These representational differences are independent of linguistic meaning but can systematically distort evaluation metrics that normalize by bytes or characters.

\subsection{Crosslinguistic Evaluation}
The most widely adopted approach relies on negative log-likelihood (NLL) computed over semantically equivalent sequences across languages. Within this framework, studies differ in how they normalize the raw NLL scores. \citet{cotterell-etal-2018-languages} and \citet{mielke-etal-2019-kind} normalize sentence-level surprisals by the number of characters of a specific language (usually English), using the resulting scores to rank languages by difficulty. \citet{limisiewicz-etal-2024-myte} adopts a similar approach but normalizes by the number of bytes instead. \citet{wan2022fairness} simply aggregates sentence-level NLL over an entire development set without normalization. A more structured alternative is proposed by \citet{tsvetkov-kipnis-2024-information}, who introduce Information Parity (IP), defined as the ratio of English NLL to the NLL of another language over parallel text.

Other work evaluates multilingual models using metrics that operate at different granularities. \citet{shliazhko-etal-2024-mgpt} report perplexity scores across languages, while \citet{blevins-etal-2022-analyzing} argue for bits per character (BPC) as a more language-neutral unit. Byte-level language models further shift toward bits per byte (BPB; \citealp{xue2022byt5,zhang-xu-2022-byte}), which \citet{shani2026roots} suggest may yield fairer comparisons when scores are scaled to a shared unit. Despite the apparent justification for each of these metrics, \citet{poelman2026form} demonstrate that applying different metrics to the same model on the same data can produce conflicting language rankings. This highlights that metric choice is not merely a technical detail but a substantive decision with interpretive consequences. This work offers concrete empirical guidance for metric selection in crosslingual evaluation and contributes to ongoing efforts toward fair multilingual NLP.


\section{Existing Intrinsic Metrics \& Their Potential Confounds}
\label{metric_analysis}

\newcommand{\mcolor}{NavyBlue}
\newcommand{\token}{\textcolor{\mcolor}{\ensuremath{t}}\xspace}
\newcommand{\tokens}{\textcolor{\mcolor}{\ensuremath{\textbf{t}}}\xspace}
\newcommand{\idx}{\textcolor{\mcolor}{\ensuremath{i}}\xspace}
\newcommand{\currtok}{\textcolor{\mcolor}{\ensuremath{\token_{\idx}}}\xspace}
\newcommand{\prevtoks}{\textcolor{\mcolor}{\ensuremath{\tokens_{<\idx}}}\xspace}
\newcommand{\sequence}{\textcolor{black}{\ensuremath{\mathbf{s}}}\xspace}
\newcommand{\numbytes}{\textcolor{orange}{\ensuremath{b}}\xspace}
\newcommand{\numchars}{\textcolor{ForestGreen}{\ensuremath{c}}\xspace}
\newcommand{\numtoks}{\textcolor[HTML]{e7298a}{\ensuremath{n}}\xspace}

\newcommand{\condp}{P(\currtok \mid \prevtoks)}

In this section, we review mainstream probability-based metrics used in language modeling and evaluate their suitability for crosslinguistic and cross-model comparison. Let \currtok be the $\idx^{th}$ token in sequence \sequence which has \numbytes UTF-8 bytes, \numchars characters, and can be tokenized into \numtoks tokens.

\paragraph{Bits per Byte}
BPB is proposed by \citet{gao2020pile} because of ``its invariance to different tokenization schemes and the ambiguity of measuring characters in Unicode.'' 
%
%
%
%
\begin{equation}
\mathrm{BPB}(\sequence) = {-\frac{1}{\numbytes} \sum_{i=1}^{\numtoks} \log_2 (\condp)}
\end{equation}

It has been used in cross-model evaluations \citep[e.g.,][]{rae2021scaling} for the same language. However, we argue that BPB introduces bias when comparing across languages due to differences in orthographic encoding. The same semantic content written in different scripts may require different numbers of UTF bytes (i.e., different \textit{byte premiums}). For example, \citet{arnett-etal-2024-bit} report that characters in scripts such as Khmer typically require three bytes per character (excluding diacritics), whereas Latin characters require only one byte. As a result, if two models assign similar probabilities to semantically equivalent sequences in two languages, the language with the higher byte premium will necessarily obtain a lower BPB score because the total surprisal $\sum_{i=1}^{\numtoks} \log_2 (\condp)$ is normalized by a larger number of bytes, \numbytes.

\paragraph{Bits per Character} 
Bits per Character (BPC) measures the average sequence NLL per character: 
\begin{equation}
\mathrm{BPC}(\sequence) = {-\frac{1}{\numchars} \sum_{i=1}^{\numtoks} \log_2 (\condp)}
\end{equation}

It is mainly used in evaluations of character-level models \citep[e.g.,][]{al2019character}. While it mitigates tokenization bias, it may still introduce orthographic bias like BPB. For example, Chinese translations are usually represented with fewer characters than English and thus it has a lower length premium. Normalizing over the number of characters means that Chinese will almost always have a higher BPC than English, as \textit{\numchars} is smaller.


\paragraph{Mean Reciprocal Rank} 
MRR \citep{limisiewicz-etal-2023-tokenization} evaluates models on the rank of the next token rather than its probability, making it probability-agnostic. However, since ranking is computed over the entire vocabulary, a larger vocabulary raises the upper bound of possible ranks, such that models with larger vocabularies tend to achieve lower MRR regardless of actual performance.
\begin{equation}
\mathrm{MRR}(\sequence) = \frac{1}{\numtoks} \sum_{t=1}^{\numtoks} \frac{1}{\mathrm{rank}(\currtok, P(\cdot \mid \prevtoks))}
\end{equation}

\paragraph{Perplexity}
Held-out perplexity measures the average number of nats (or bits) required to predict tokens in a corpus. It has been the default metric for evaluating language models, with lower values typically indicating better performance \citep[e.g.,][]{brown2020language}:
\begin{equation}
\mathrm{PPL}(\sequence) = \mathrm{exp}({-\frac{1}{\numtoks} \sum_{i=1}^{\numtoks} \ln (\condp)})
\end{equation}

Some previous studies report perplexity as the measure to compare models with different tokenizers and training data \citep[e.g.,][]{thakur2025art}. Because perplexity is normalized by the number of tokens, it is highly sensitive to how the input is tokenized. Consider a language model trained on Language \textit{A} and evaluated on a parallel corpus containing both Language \textit{A} and Language \textit{B}, where the two languages use different scripts. Since the model has never been exposed to Language \textit{B}'s script, its tokenizer will decompose Language \textit{B}'s text into smaller units (down to bytes under standard BPE) and produce longer token sequences. As the denominator increases, the normalized log-likelihood decreases, artificially deflating perplexity for Language \textit{B} despite the model having no genuine knowledge of it.\footnote{We present a case study using \texttt{GPT2} \citep{radford2019language} in Appendix~\ref{app:case_study} to illustrate this point.}


\paragraph{Token-level Negative log-likelihood}
Token-level NLL (Token NLL) measures the NLL of the sequence normalized by the number of tokens: 
\begin{equation}
\text{Token NLL}(\sequence) = {-\frac{1}{\numtoks}\sum_{i=1}^{\numtoks} \ln (\condp)} 
\end{equation}

Token NLL shares the same tokenization sensitivity as perplexity, as discussed above. 

\paragraph{Sentence-level Negative log-likelihood} \label{sec:metrics}
Sentence-level NLL (Sent-NLL) is the sum of the NLL (surprisal) per token over a text sequence (See \ref{nll}). Note that the sequence need not be a linguistic sentence; it can be any unit of parallel text that conveys equivalent meaning across languages, such as a paragraph or document.

\begin{equation}\label{nll}
\text{Sent-NLL}(\sequence) = {-\sum_{i=1}^{\numtoks} \ln (\condp)}  
\end{equation}

Because Sent-NLL sums the surprisal over the entire sequence without normalizing by token count, it measures the model’s surprisal for semantically equivalent content expressed in different languages. Therefore, it is not directly affected by how a sequence is segmented. For this reason, Sent-NLL of parallel sentences has been more widely adopted in crosslinguistic comparisons \citep{lin2024mala, costa2022no, chang2024goldfish}, and we argue that it provides the most meaningful basis for crosslingual evaluation among the metrics considered here. \looseness=-1

\paragraph{Other metrics} There are other metrics such as Information Parity \citep[IP; ][]{tsvetkov-kipnis-2024-information} and Bits per English Character \citep[BPEC;][]{cotterell-etal-2018-languages}. As they are transformations of Sent-NLL covered here, they should align with the results of Sent-NLL. 

\begin{figure*}[t!]
    \centering
    \includegraphics[width=\linewidth]{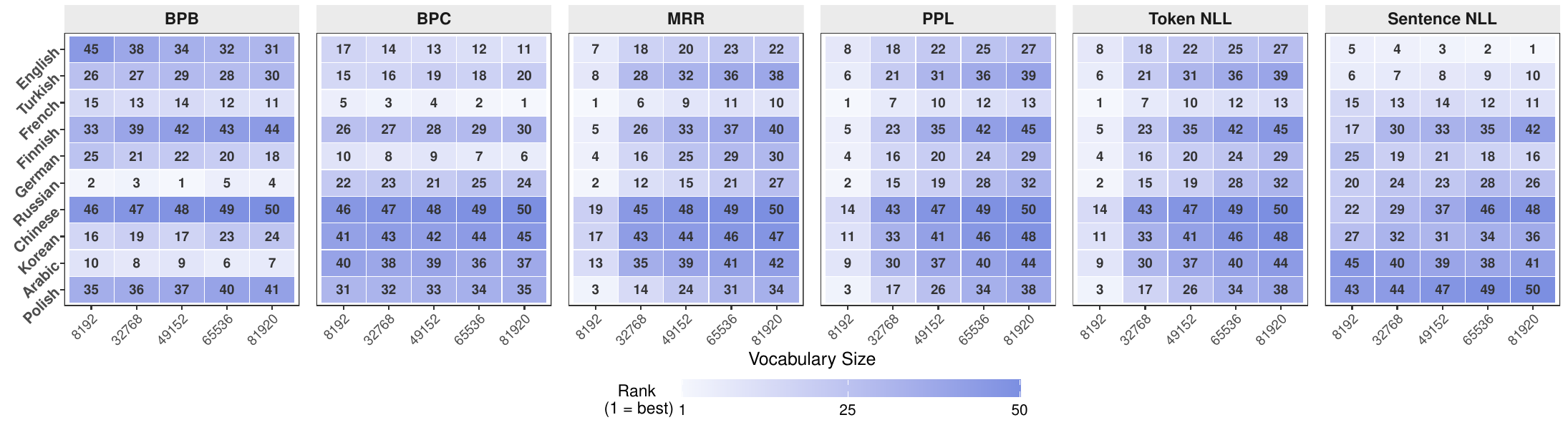}
   \caption{Language model rankings across intrinsic metrics. Each cell shows the rank of a language-vocabulary size combination among all 50 settings (10 languages * 5 vocabulary sizes) within each metric. Lighter colors indicate better performance (rank 1 = best).}
    \label{fig:metrics}
\end{figure*}

\paragraph{Summary} As summarized in Figure~\ref{fig:metric_suggesetion}, normalization over tokens, characters, or bytes introduces systematic biases and is unsuitable for cross-tokenizer or cross-lingual comparison.  BPB and BPC are appropriate for cross-tokenizer comparison within the same language, but not cross-lingually, as different writing systems yield different byte and character counts. Perplexity and Token-NLL are unsuitable for cross-lingual or cross-tokenizer comparisons because different tokenizers can segment the same content into different numbers of tokens across languages. MRR requires constant vocabulary size across models. Only Sent-NLL over semantically equivalent content is suitable for both cross-model and cross-lingual comparison, as it normalizes by meaning rather than representation units.

\section{Experiment 1: All metrics except for Sent-NLL are biased in different ways}
\label{exp1}

To ensure that observed metric differences reflect the metrics themselves rather than model-specific confounds,\footnote{For example, a multilingual model trained predominantly on Chinese may assign lower BPC to Chinese text, masking BPC's inherent bias against logographic writing systems: Chinese characters encode more information per character than alphabetic scripts, artificially shortening sentences and improving BPC scores.} we use monolingual models trained on parallel corpora with varying vocabulary sizes.

\paragraph{Experiment Setup.}

We use \texttt{GPT2} models from concurrent work \citep{yang2026}: monolingual models, each trained exclusively on one language's portion of a parallel corpus. This design ensures that cross-linguistic differences are attributable to the languages themselves rather than to shared parameters or training data imbalances.

Two model sizes are used. \texttt{GPT2-small} models are trained on \textsc{Parallel-10}, a ten-language parallel corpus spanning six language families ($\sim$20M words per language),\footnote{Arabic, Chinese, English, French, German, Finnish, Polish, Russian, Korean, and Turkish.} with five vocabulary sizes,\footnote{8192, 32768, 49152, 65536, 81920.} yielding 50 models in total. \texttt{GPT2-medium} models are trained on \textsc{Parallel-3}, a trilingual corpus (Chinese, English, Arabic; $\sim$400M words per language) with two vocabulary sizes (32k and 65k), yielding 6 models, allowing us to test whether findings hold at scale. The details of the training corpora can be found in \Cref{sec:parallel-corpus}.

\begin{table}[t]
\small
    \centering
    \begin{tabular}{l|lcc}
    \toprule
    Metric & Potential Confounds  &  $\rho$ & $p$\\
    \midrule
    BPC  & \#characters & -0.81 & 0.0082 \\
    BPB & \#bytes & -0.94 & <0.001 \\
    PPL & CTC& -0.93 & <0.001 \\
    MRR & CTC &  0.89 & 0.001 \\
    Token NLL & CTC&  -0.93 & <0.001\\
    \midrule
     Sent-NLL & \#characters &  -0.38& 0.28\\
     Sent-NLL & \#bytes &  0.14& 0.71\\
    Sent-NLL & CTC &  -0.19 & 0.19\\
    \bottomrule
    \end{tabular}
    \caption{Correlation Analysis between metrics and potential confounds. PPL is scaled by log10.}
    \label{tab:corr}
\end{table}
\paragraph{Evaluation.}
\label{sec:evaluation}
We evaluate the models using the metrics discussed in Section~\ref{metric_analysis} on three parallel data sources in the corresponding languages: FLORES-200 \citep{flores}, a multi-parallel machine translation benchmark covering more than 200 languages;  the in-domain test split of our training corpus (40k sentences, see Table~\ref{tab:data-splits}); and 1,000 sentences from the Parallel Universal Dependencies treebanks \citep[PUD;][]{zeman-etal-2017-conll}.

A fair metric should be less sensitive to encoding and tokenization choices. Based on the discussion in \Cref{metric_analysis}, we expect language rankings based on BPB and BPC to be largely predictable from the number of bytes and characters, respectively, and rankings based on Token-NLL, PPL, and MRR to be largely predictable from the number of tokens (i.e., CTC). In contrast, we do not expect language rankings based on Sent-NLL to exhibit such relationships. We examine these predictions in Figure~\ref{fig:metrics} and quantify them using Spearman’s rank correlation. \looseness=-1

\paragraph{Results}
\label{exp1-result}

Figure~\ref{fig:metrics} reports rankings across 10 languages and 5 vocabulary sizes on FLORES-200 for \texttt{GPT2-small} (results on other datasets are in Appendix~\ref{parallel10}). Results for \texttt{GPT2-medium} can be found in \Cref{gpt-medium}. Different metrics produce strikingly different crosslingual rankings. \citet{poelman2026form} attribute this to the general unreliability of probability-based metrics. While this may be true, in part, we argue that the inconsistencies are attributable to specific, identifiable biases in the normalization unit of each metric. 

We observe the expected patterns in Figure~\ref{fig:metrics}. BPB and BPC consistently rank Chinese last, as Chinese has shorter character and byte sequences. MRR, PPL, and Token NLL rankings are dominated by vocabulary size: smaller vocabularies yield better rankings regardless of language. This ranking arises for different reasons. For PPL and Token NLL, a tokenizer with a smaller vocabulary size produces more tokens (i.e., a higher CTC). As these metrics are averaged over tokens, a higher CTC can result in lower PPL and Token NLL (i.e., better performance), even when the underlying modeling quality is unchanged. For MRR, a smaller vocabulary size reduces the upper bound of possible ranks, which can make models with smaller vocabulary sizes appear to perform better. The same pattern holds in our scaling experiments. These results suggest that rankings under surface-normalized metrics are largely predictable from engineering choices alone, rather than reflecting genuine differences in language modeling quality. In contrast, Sent-NLL shows less sensitivity to either encoding or vocabulary size.

\begin{figure*}[t]
    \centering
    \includegraphics[width=0.93\linewidth]{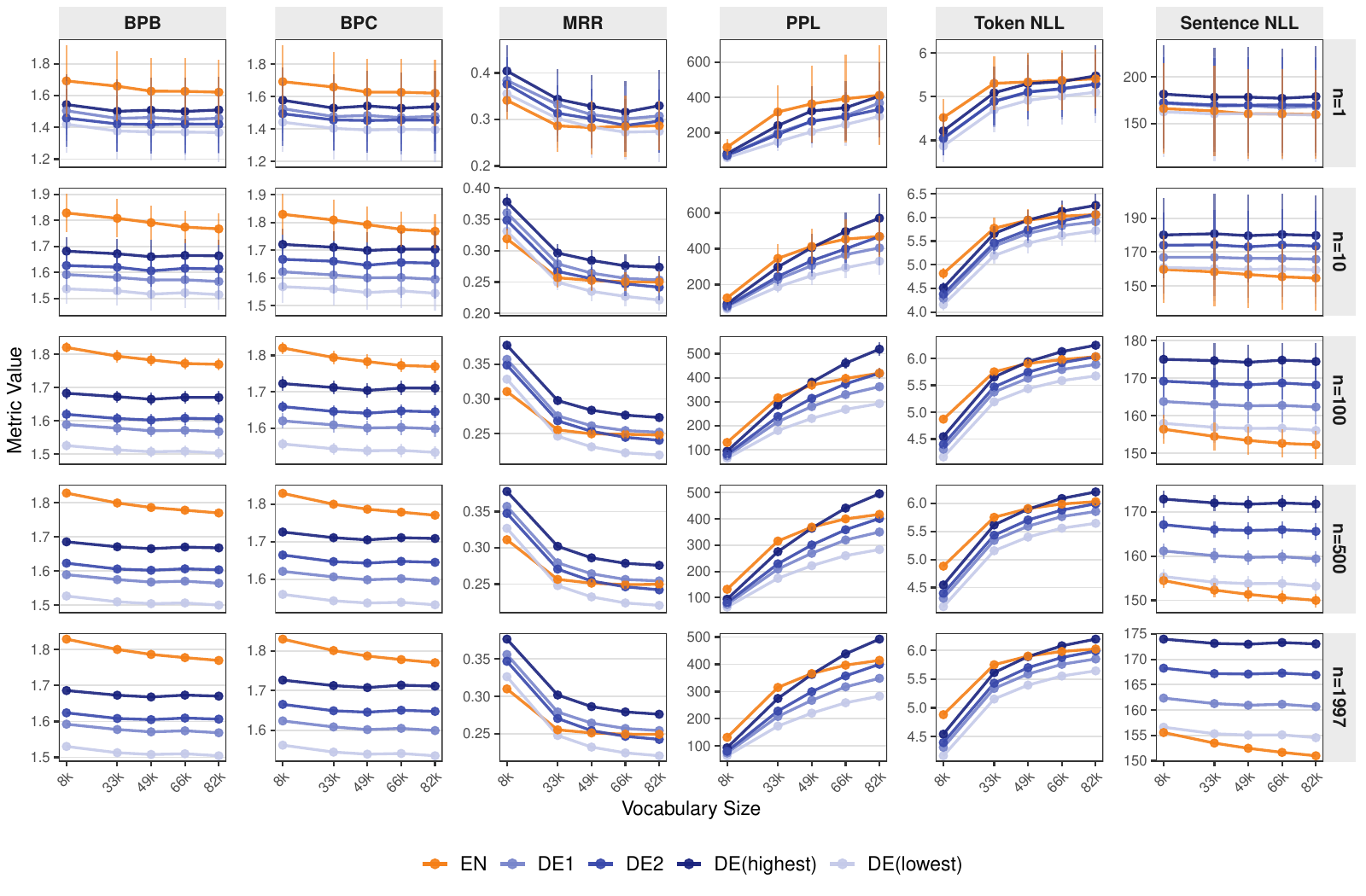}
    \caption{Each row shows results for a different sample size. Error bars represent 95\% CI across 15 random seeds.}
    \label{fig:sample_size}
\end{figure*}

To substantiate this claim, we conduct Spearman’s rank correlation tests between each metric and the corresponding potential bias factors discussed in Section~\ref{metric_analysis}. For BPC and BPB, the number of characters and bytes for a given language remains constant regardless of tokenization, so we average model performance across tokenizers before computing correlations. For other metrics, the potential confound is vocabulary size, which we operationalize as CTC and use in our analysis.\footnote{Because for a given corpus tokenized by the same tokenization algorithm, vocabulary size and CTC are in a one-to-one correspondence, we use CTC for consistency in our following analysis.} As CTC varies across tokenization settings because different tokenizers produce different numbers of tokens for the same language, we therefore compute correlations between each model’s metric value and its corresponding CTC directly, without averaging across tokenizers. For Sent-NLL, we compute correlations with all three factors. 

Both the correlation coefficient $\rho$ and p-value $p$ of the correlation analysis are reported in Table~\ref{tab:corr}. We find that the potential confounds are strongly and significantly correlated with language rankings for all metrics ($\rho$>0.8, $p$<0.01) except Sent-NLL, which shows no significant correlation with sequence length measured in bytes, characters, or tokens.


\section{Experiment 2: Sent-NLL is more robust for translation alternatives }
\label{exp2}

One potential objection to Sent-NLL is that different translations of the same text may yield different probability distributions, leading to inconsistent crosslingual rankings \citep{poelman2026form}. However, \citet{poelman2026form} demonstrate this instability using both translations and paraphrases within the same language rather than translations only. Because paraphrases allow more flexible wording as long as the gist of the meaning is kept \citep{neusner1986translation,bhagat-hovy-2013-squibs}, they are more likely to encode different amounts of information. In this case, their findings speak to within-language variation rather than the validity of crosslingual comparison over parallel translations. We provide additional discussion about this assumption in Section \ref{sec:paraphrase}.

\paragraph{Experiment Setup.}  To test each metric's sensitivity to meaning-equivalent translations, we re-evaluate \citeauthor{poelman2026form}'s (\citeyear{poelman2026form}) experiment using translations only and additionally evaluate Sent-NLL. Specifically, we use the English-German translation pairs from WMT2019 \citep{barrault-etal-2019-findings} with an additional translation reference source created independently from \citet{freitag-etal-2020-human,freitag-etal-2020-bleu}, denoted as DE$1$ and DE$2$ separately. Following \citet{poelman2026form}, for these two alternative German translations, we construct two derived datasets: DE$_{lowest}$ and DE$_{highest}$, consisting of sentences that receive the lower or higher metric value in each pairwise comparison, respectively. \looseness=-1

We use the same sets of models trained on English and German in Experiment 1 and compare Sent-NLL for English and German across the five data sources: EN (English source text), DE$1$ (German translation), DE$2$ (German alternative translation), DE$_{lowest}$, and DE$_{highest}$.

A fair metric should be robust to translation alternatives. That is, for a given source text, evaluating models on different translations should yield consistent language rankings. Based on the analysis in \Cref{metric_analysis}, we predict that Sent-NLL will yield consistent rankings across translation alternatives, whereas the other metrics will be more sensitive to translation variation.

One possible objection is that even for semantically equivalent translations, sentence-level rankings may not align across languages, potentially affecting aggregated language rankings \citep{poelman-etal-2025-confounding}.  We argue that such sentence-level mismatches introduce noise, and that their impact diminishes when averaged over a sufficiently large number of parallel sentences. To test this, we conduct a follow-up experiment under the same setting while varying the number of samples from 1 to the full dataset. For each sample size, we repeat the experiment with 15 random seeds. 

\paragraph{Results.} Figure~\ref{fig:sample_size} shows the results of the replication experiment (last row) and the sampling experiment (first three rows).  

The replication experiment reveals three major findings. First, metrics including MRR, PPL, and Token NLL vary monotonically by vocabulary size, which is consistent with our findings in Section~\ref{exp1}. Second, for these metrics, even with the same vocab size setting, using different translations in German yields different language rankings between English and German. This pattern is especially salient when vocab size $\geq$ 33k. This suggests that these metrics are more sensitive to surface-level differences. Finally, we find that Sent-NLL produces consistent language rankings across translation variants (i.e., English scores lower than all four German datasets). We observe the same consistency for BPB and BPC, but with different rankings. However, as we show in Section~\ref{exp1}, this apparent stability is driven in part by language-specific encoding factors.


 \begin{figure*}[t]
    \centering
    \includegraphics[width=0.93\linewidth]{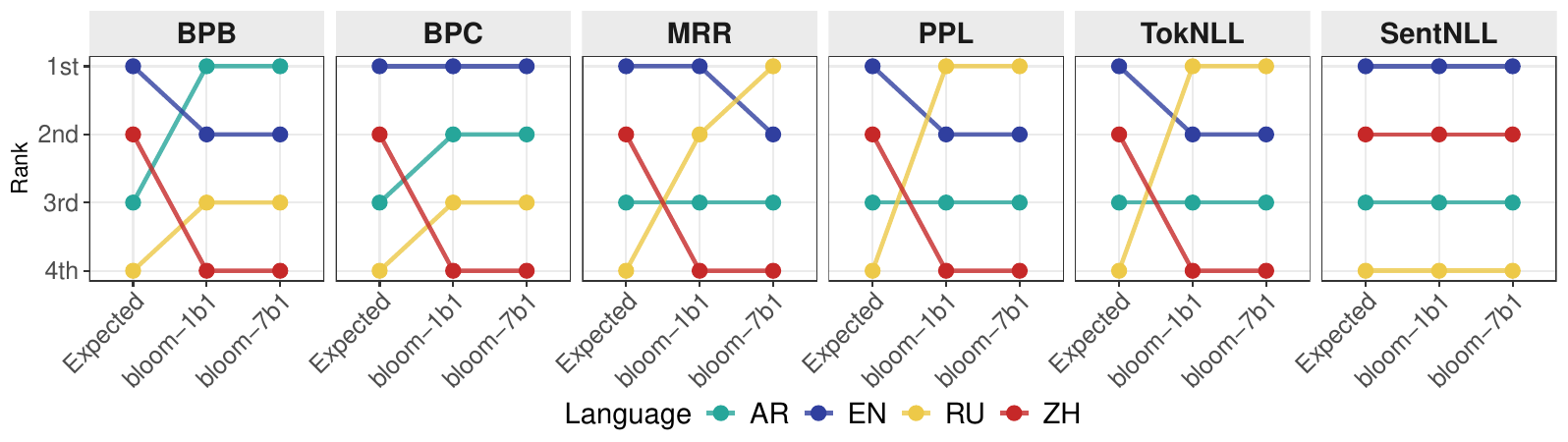}
    \caption{Expected ranking and actual ranking across 6 metrics for \texttt{BLOOM-7b1} and \texttt{BLOOM-1b1}. }
    \label{fig:multilingual-comp}
\end{figure*}

Regarding the sampling experiment, we find that when the sample size is small (1–100), the rankings are indeed unstable. However, they become stable once the sample size reaches around 500. We note that all other experiments in this paper use sample sizes greater than 500.

\section{Experiment 3: Sent-NLL generalizes to multilingual language models}
\label{exp3}
The metric biases we identify are mathematical in nature and therefore model-agnostic: they arise from how metrics normalize probability, not from properties of any particular model. 

\paragraph{Experiment Setup.} To examine this, we evaluate nine multilingual LLMs on four languages (Arabic, English, Russian, Chinese) from FLORES-200. We select two groups of models based on different criteria. First, \texttt{BLOOM} \citep{workshop2022bloom}, \texttt{mGPT} \citep{shliazhko-etal-2024-mgpt}, and \texttt{XGLM} \citep{lin-etal-2022-shot} are selected because their training data composition is well-documented, allowing us to derive expected language rankings. Second, \texttt{Teuken} \citep{ali2024teuken}, \texttt{Qwen} \citep{yang2025qwen3}, and \texttt{Llama} \citep{grattafiori2024llama} are included due to their wide usage in multilingual settings. Training data composition is reported in Table~\ref{tab:data_comp}. For the first group, we compare expected language rankings, derived from training data proportions under the assumption that more data leads to better performance \citep{hoffmann2022training}, against actual metric rankings. For the second group, whose full training data composition is undisclosed, we verify whether the dominant language reported in the technical documentation achieves the best ranking under each metric. A reliable metric should produce rankings consistent with these expectations.

\paragraph{Results.} 

Results for \texttt{BLOOM-7B} and \texttt{BLOOM-1B} are shown in Figure~\ref{fig:multilingual-comp} and confirm our predictions. The leftmost column shows the expected ranking based on training data composition; the right two columns show actual rankings under each metric. BPB and BPC are distorted by script differences: Chinese consistently ranks worst despite comprising the second largest portion of \texttt{BLOOM}'s training data.  PPL and Token-NLL are distorted by tokenization: Russian, which receives little training data in \texttt{BLOOM}, paradoxically ranks best because unseen languages are over-segmented into more tokens, artificially deflating token-normalized scores. Only Sent-NLL produces rankings that reflect actual training data composition. We find the same pattern with other LLMs; raw results for all models are reported in Figure~\ref{fig:raw_results}, and ranking results for \texttt{XGLM-4.5B} and \texttt{mGPT} in  Figure~\ref{fig:GLM}.

\section{Considerations for Downstream Tasks} 

While intrinsic metrics are our primary focus, downstream task evaluations remain widely used in practice. However, fair crosslingual comparison using downstream tasks is far from straightforward. Below, we discuss several considerations for fair comparisons across languages using downstream evaluations.

\paragraph{Translated Benchmarks.} Some translated tasks may enable comparable evaluation across languages. For example, a math benchmark translated into another language may allow a researcher to make an inference about the relative performance on math tasks in two languages. Math and reasoning benchmarks are less likely to be influenced by translation quality and are more language-agnostic \citep{wu2025bitter}. This relies on sufficiently high translation quality, however. Machine translated benchmarks without human validation may contain translation artifacts and introduce noise into the evaluation \citep{singh-etal-2025-global}. High-quality benchmarks created through professional human translation, e.g. MMMLU \citep{openai-mmmlu} or FLORES \citep{flores} and derivatives like Belebele \citep{bandarkar-etal-2024-belebele} or SIB-200 \citep{adelani-etal-2024-sib}, are less likely to suffer from these issues. \looseness=-1

\paragraph{Culture-specific tasks.}
Another consideration is cultural specificity. Even when translated faithfully, some benchmarks may not be appropriate for comparable evaluation if their content is specific to a particular linguistic or cultural context. 
Localized benchmarks, e.g. IndoMMLU (Indonesian; \citealp{koto-etal-2023-large}), CMMLU (Chinese; \citealp{li-etal-2024-cmmlu}), Turkish MMLU \citep{yuksel-etal-2024-turkishmmlu}, Arabic MMLU \citep{koto-etal-2024-arabicmmlu}, and KMMLU (Korean; \citealp{son-etal-2025-kmmlu}) evaluate not only target-language performance but also cultural knowledge. Localized benchmarks such as these correlate more strongly with human preferences than translated benchmarks \citep{wu2025bitter}. But achieving the same score on two localized MMLU-style benchmarks does not signal comparable performance, as the questions are not difficulty-matched or otherwise controlled. 

\paragraph{Language-specific tasks.}
Language-specific tasks, such as linguistic knowledge benchmarks like BLiMP \citep{warstadt2020blimp} and MultiBLiMP \citep{jumelet2026multiblimp}, may not be comparable across languages for other reasons. First, even benchmarks which evaluate the same relatively narrow tasks, such as subject-verb agreement in MultiBLiMP \citep{jumelet2026multiblimp}, are based on non-parallel data. The items for each language differ in quantity and quality. The number of items in MultiBLiMP per language ranges from 7 (Gujarati) to 4615 (Old Russian). For languages with very few items, like Gujarati, MultiBLiMP scores may be less reliable than those with much larger numbers of samples. Additionally, as MultiBLiMP is based on UD Treebanks \citep{de-marneffe-etal-2021-universal,nivre-etal-2020-universal}, which vary in their annotation schema. These inconsistencies may lead to different conclusions about relative linguistic complexity, for example \citep{rodriguez-lopez-2025-beyond}. Both of these issues can be addressed through the use of balanced parallel data with consistent annotation schema. \looseness=-1

These benchmarks are also sensitive to crosslinguistic differences. In the development of CLAMS \citep{mueller-etal-2020-cross}, the authors discuss how differences in pronoun dropping, grammatical person distinction, and distribution of negative polarity items shaped the types of tests they could use in the comparison of agreement across languages.

Subject-verb agreement is inherently different across languages. For some languages, the task is trivial as there is no agreement and the verb form is always the same (e.g. Chinese and Tagalog; \citealp[]{chao1968grammar, reyes1969some}). Other languages have more complex agreement patterns due to higher number of distinctions between number (singular, dual, and plural number, e.g. Arabic; \citealp[]{bettega2022gender}), gender/noun class (about 10 noun classes in Swahili; \citealp{moxley1998semantic}), verb class (at least 4 verb classes in Georgian; \citealp[]{makharoblidze2012georgian}), gender (Tamil; \citealp[]{venkatesan2024gender}).
It remains an open question how to fairly compare linguistic performance across languages, when the complexity of the phenomenon differs.








\section{Discussion \& Conclusion}

As language models are increasingly deployed beyond English, fair crosslingual evaluation is a necessity. How we evaluate models determines how we interpret results and allocate resources. Prior work has noted that normalized metrics may be unreliable for cross-lingual comparison in monolingual or multilingual settings \citep{cotterell-etal-2018-languages, mielke-etal-2019-kind}, but these observations have remained largely theoretical or have been based on intuition. Problematic metrics continue to be used in different studies \citep{shliazhko-etal-2024-mgpt, arnett-bergen-2025-language, thakur2025art} or explicitly advocated in recent proposals \citep{shani2026roots}. Our contribution is to provide the first systematic empirical evidence across multiple metrics, languages, and tokenizer settings. Such empirical grounding matters precisely because intuition alone has not been sufficient to change community norms.\looseness=-1

It is important to note that Sent-NLL is not entirely free from confounds. However, we argue that any remaining sensitivity reflects inherent linguistic complexity rather than arbitrary engineering choices, which is a qualitatively different kind of bias that is arguably unavoidable in any crosslingual comparison. Our proposal implicitly assumes that all languages encode roughly comparable amounts of information, an assumption the literature both supports and challenges \citep{koplenig2017statistical,crystal_equality_2010,fenk2014complexity,bentz2023complexity}. Even if languages systematically differ in information density, such differences reflect intrinsic linguistic properties rather than metric artifacts. In this sense, Sent-NLL may reflect both how well a model has learned a language and the intrinsic information density of that language. Therefore, even if the linguistic equi-complexity hypothesis, i.e., all natural languages are equally complex, is falsified, Sent-NLL may not be a perfect measure of model performance in isolation, but it remains the least biased option among commonly used intrinsic metrics for crosslingual comparison. Any remaining confound due to linguistic complexity should be viewed as a limitation shared by all crosslingual evaluation rather than a systematic bias of the metric itself.

Finally, our results show that Sent-NLL rankings show more variability across datasets than normalized metrics (Figures~\ref{fig:metrics}, \ref{fig:metrics-parallel10}, \ref{fig:metrics-pud}). We argue this reflects a strength rather than a weakness. The apparent stability of metrics such as BPB, BPC, PPL, and Token-NLL is artificial: their rankings are largely determined by engineering choices -- script properties, tokenization granularity, and vocabulary size -- rather than genuine differences in language modeling quality. Sent-NLL, by contrast, is sensitive to actual test data content, meaning its rankings reflect real differences in how well a model has learned each language. Variability across datasets is therefore expected and desirable.

\section*{Limitations}
Our evaluation only covers 10 languages due to limited parallel data. A broader language sample would strengthen generalizability, although the significant correlations observed for all other metrics under the same sample size suggest that statistical power is sufficient to detect meaningful relationships.

In the paraphrase experiments, only English and German translation pairs are tested due to the limited translation resources. Extending this analysis to additional bilingual pairs with multiple translation references would further strengthen the conclusions.

More broadly, our approach relies on high-quality parallel sentences. Although datasets such as FLORES provide relatively broad coverage, many languages still lack reliable parallel data, which limits the applicability of this evaluation framework.

\section*{Acknowledgments}
We thank Tyler A. Chang and anonymous reviewers for their helpful feedback. We also thank CoreWeave for providing the GPU resources used for model training and evaluation.

\bibliography{custom}
\appendix

\section{Parallel Multilingual Corpus}
\label{sec:parallel-corpus}
The training data sources and split can be found in Table~\ref{tab:parallel_sources} and Table~\ref{tab:data-splits}.
We report the typological features for each language in \Cref{typology}.

\begin{table*}[t]
\small
\centering
\begin{adjustbox}{max width=\textwidth}
\begin{tabular}{llrrrrrr}
\toprule
\multirow{2}{*}{Source} &\multirow{2}{*}{Description} & \multicolumn{2}{c}{parallel-10} & \multicolumn{2}{c}{parallel-3} \\
 \cmidrule(lr){3-4}\cmidrule(lr){5-6}
& & Words & Sents & Words & Sents \\
\midrule
OpenSubtitles \citep{lison-tiedemann-2016-opensubtitles2016} & Movie/TV show subtitles & 17.0M & 2.0M & 64.9M & 8.1M \\
Bible \citep{christodouloupoulos2015massively}      & Bible text  & 1.5M  & 57K  & 1.5M & 61.2K \\
WikiTitle    & Titles of Wikipedia articles & 167K  & 42K  & 848.2K &218.3K  \\
Neulab-Ted   & Ted talks & 356K  & 19K  & 3.7M & 201.2K \\
KDE4  & KDE4 localization files & 117K  & 13K  & 280.7K & 30.0K \\
Ted2020 \citep{reimers-gurevych-2020-making}    & Ted talks  & 347K  & 19K  & 3.1M &171.0K\\
QED       &  Subtitles for educational videos & 10K   & 0.6K & 73.5K & 5.0K \\
Ubuntu   &   Ubuntu localization files   & <1K  & <100   & 2.5K &  <1K\\
Tatoeba   & Translated sentences from Tatoeba    & <1K  & <100   &20.2K  & 2.9K \\
GNOME    &   GNOME localization files   & <1K  & <100   & <1K &  <100\\
\midrule
UNPC \citep{ziemski-etal-2016-united}      & United Nation documents   &   NA    &   NA   & 306.0M & 10.7M \\
GlobalVoices & News report &   NA    &  NA    & 156.0K &  7.7K\\
NewsCommentary \citep{barrault-etal-2019-findings} & news comments & NA    &  NA    & 2.5M & 55.6K \\
Ted2013       & Ted talks &   NA    &  NA    & 983.9K&  171.0K\\
Tanzil       & Quran translations &   NA    & NA   & 14.2K & 1.7K\\
\midrule
\textit{Overall}  &      & 19.5M & 2.4M & 384.2M & 19.6M \\
\bottomrule
\end{tabular}
\end{adjustbox}
\caption{Sources of the parallel-10 and parallel-3 corpora, from \citet{yang2026}.}
\label{tab:parallel_sources}
\end{table*}

\begin{table*}[]
\small
    \centering
 \begin{tabular}{l l|rr rr rr}
\toprule
\multirow{2}{*}{Dataset} & \multirow{2}{*}{Language} &
\multicolumn{2}{c}{Train} & \multicolumn{2}{c}{Dev} & \multicolumn{2}{c}{Test} \\
\cmidrule(lr){3-4} \cmidrule(lr){5-6} \cmidrule(lr){7-8}
 & & \#sent & \#word & \#sent & \#word & \#sent & \#word \\
    \midrule
    \multirow{10}{*}{parallel-10} 
& English & 2,338,286 & 19,089,154 & 10,000 & 82,457 & 40,000 & 327,724 \\
& Arabic & 2,338,286 & 14,426,353 & 10,000 & 61,982 & 40,000 & 246,954 \\
& Chinese & 2,338,286 & 30,129,452 & 10,000 & 130,205 & 40,000 & 516,401 \\
& Turkish & 2,338,286 & 12,540,711 & 10,000 & 54,238 & 40,000 & 215,345 \\
& German & 2,338,286 & 16,544,740 & 10,000 & 71,464 & 40,000 & 283,997 \\
& Korean & 2,338,286 & 11,857,382 & 10,000 & 51,292 & 40,000 & 203,336 \\
& Polish & 2,338,286 & 13,458,472 & 10,000 & 57,932 & 40,000 & 231,262 \\
& French & 2,338,286 & 17,603,323 & 10,000 & 76,087 & 40,000 & 302,056 \\
& Russian & 2,338,286 & 14,613,449 & 10,000 & 62,951 & 40,000 & 251,018 \\
& Finnish & 2,338,286 & 11,975,617 & 10,000 & 51,698 & 40,000 & 205,766 \\
    \midrule
    \multirow{3}{*}{parallel-3} 
& English & 19,555,763 & 383,187,466 & 10,000 & 198,658 & 40,000 & 780,968 \\
& Arabic & 19,555,763 & 332,197,948 & 10,000 & 171,796 & 40,000 & 676,630 \\
& Chinese & 19,555,763 & 677,898,235 & 10,000 & 349,927 & 40,000 & 1,380,368 \\
    \bottomrule
    \end{tabular}
    \caption{Data splits for three experimental settings across ten languages. Word counts are computed using whitespace-separated tokens for all languages except Chinese, for which each character is treated as a word, from \citet{yang2026}.}
    \label{tab:data-splits}
\end{table*}

\begin{table}[!th]
    \centering
    \small
\begin{tabular}{cr}
\toprule
\multicolumn{2}{c}{Training}\\
\midrule
\texttt{train\_size} & \texttt{10M} \\
\texttt{num\_epoch} & \texttt{10}\\ \texttt{train\_amount} & 
\texttt{19M} \\
\texttt{batch\_size} & \texttt{128}\\
\texttt{context\_length} & \texttt{128}\\
\texttt{grad\_acc\_steps} & \texttt{1}\\
\texttt{weight\_decay} & \texttt{0.1}\\
\texttt{warmup\_steps} & \texttt{10\%}\\
\texttt{lr} & \texttt{5e-4}\\
\texttt{lr\_scheduler} & \texttt{linear}\\
\bottomrule
\end{tabular}
\caption{Training Hyperparameters}
\label{training_hype}
\end{table}
All the sources follow the data license from OPUS \citep{TIEDEMANN12.463}.

\section{Model Training Details}
\label{hyperparameter}
The training hyperparameters are reported in Table~\ref{training_hype}.

\section{Supplementary results for Experiment1}
\subsection{Dataset specific results for monolingual \texttt{GPT2-small}}
\label{parallel10}
\begin{figure*}[!th]
    \centering
    \includegraphics[width=\linewidth]{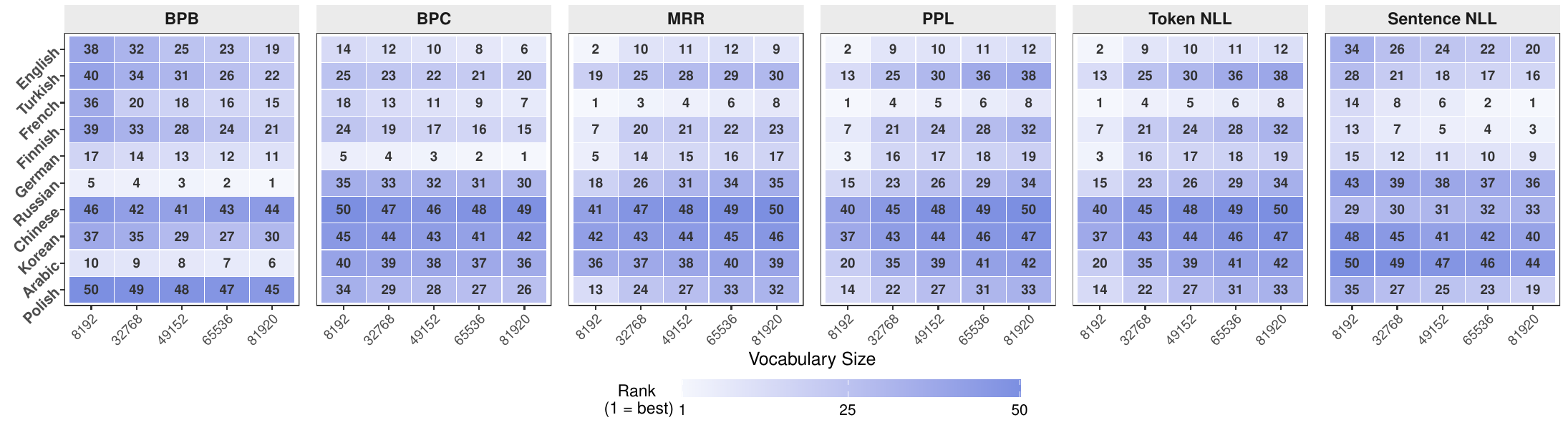}
    \caption{Language rankings across different intrinsic metrics for Parallel-10 test split}
    \label{fig:metrics-parallel10}
\end{figure*}

\begin{figure*}[!th]
    \centering
    \includegraphics[width=\linewidth]{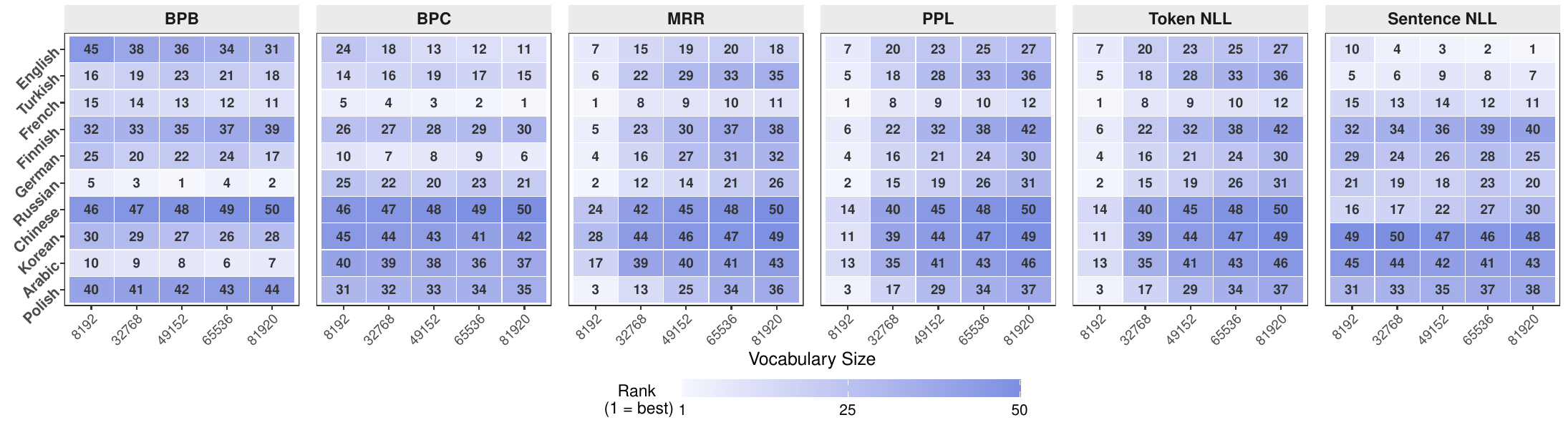}
    \caption{Language rankings across different intrinsic metrics for PUD}
    \label{fig:metrics-pud}
\end{figure*}

\begin{figure*}[ht]
    \centering
    \includegraphics[width=\linewidth]{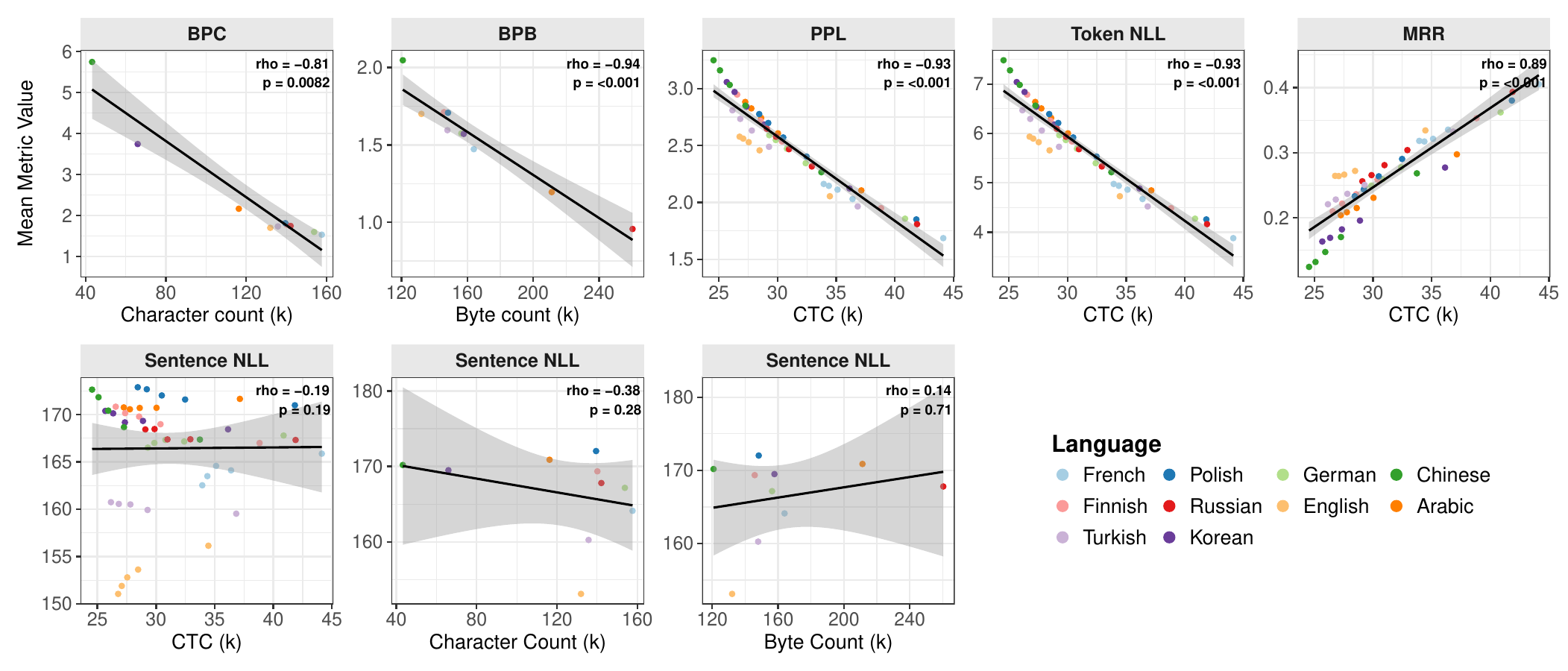}
    \caption{Spearman's Rank Correlation between the intrinsic metrics and their potential confounding factors in FLORES (perplexity is scaled by log10).}
    \label{fig:metric_correlation}
\end{figure*}

\begin{figure*}[ht]
    \centering
    \includegraphics[width=\linewidth]{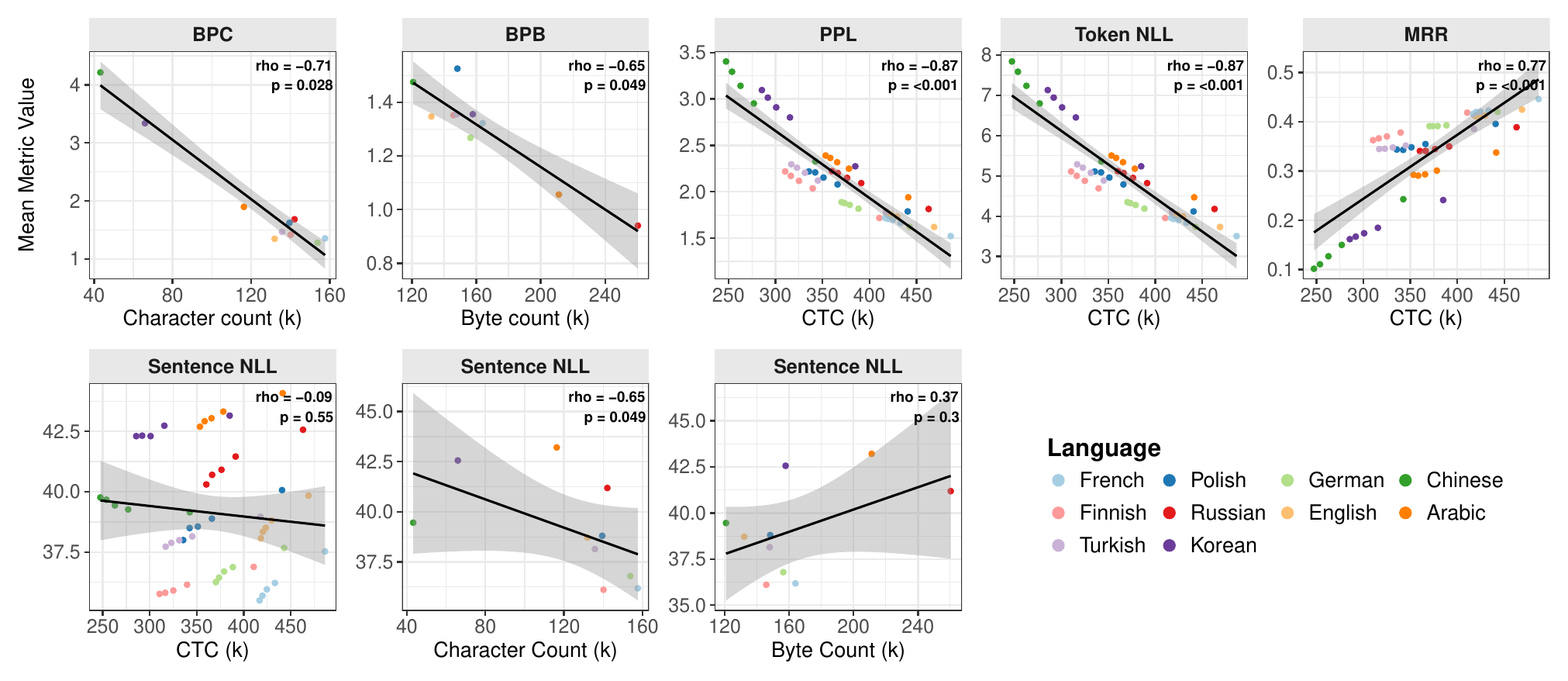}
    \caption{Spearman's Rank Correlation between the intrinsic metrics and their potential confounding factors in Parallel10 (perplexity is scaled by log10).}
    \label{fig:metric_correlation_parallel10}
\end{figure*}

\begin{figure*}[!ht]
    \centering
    \includegraphics[width=\linewidth]{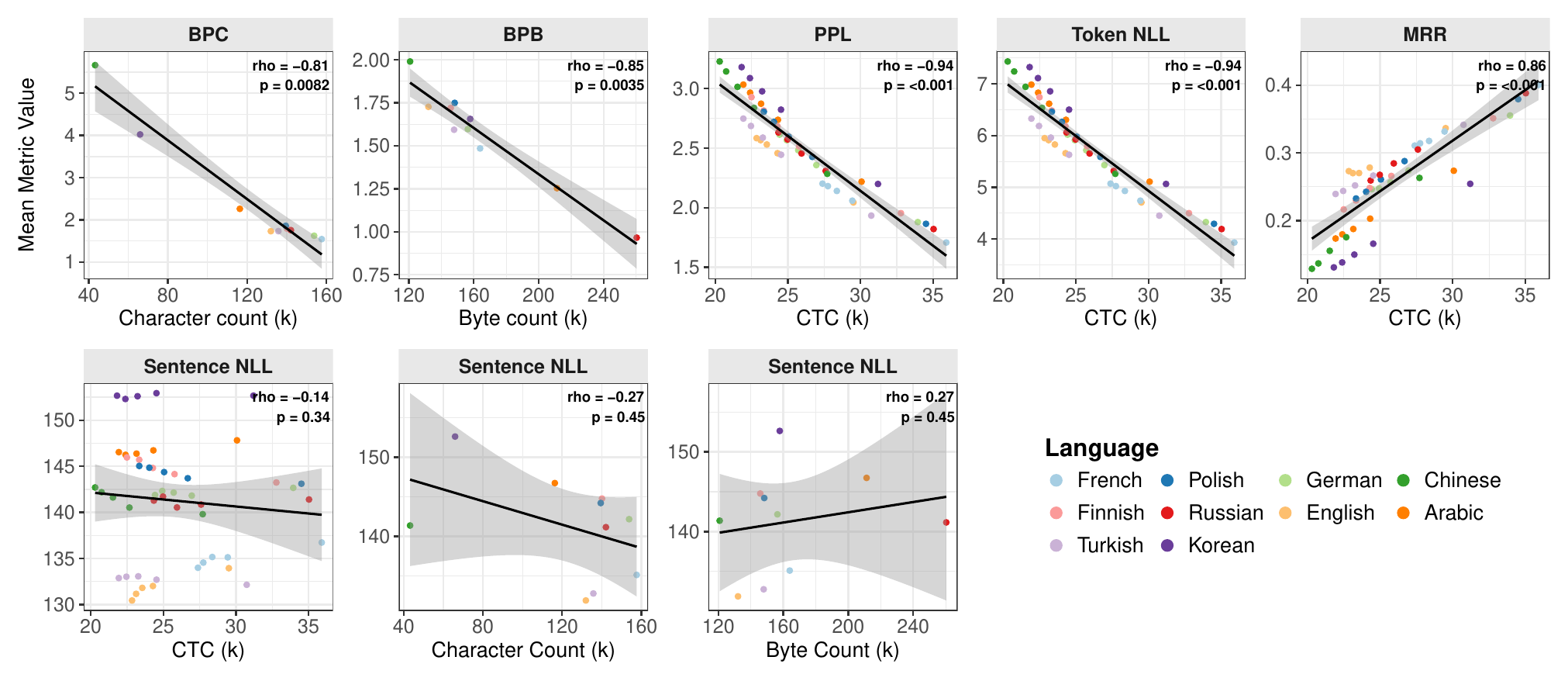}
    \caption{Spearman's Rank Correlation between the intrinsic metrics and their potential confounding factors in PUD (perplexity is scaled by log10).}
    \label{fig:metric_correlation_pud}
\end{figure*}

The crosslingual rankings for Parallel-10 test split, and PUD are shown in Figures~ \ref{fig:metrics-parallel10}, \ref{fig:metrics-pud} separately. The correlation test are visualized in Figures~\ref{fig:metric_correlation}, \ref{fig:metric_correlation_parallel10}, \ref{fig:metric_correlation_pud} separately. We observe similar patterns as discussed in the main text. In the correlation test, we find significant correlation between each metric and their potential confounds across three datasets, but this is not the case for Sent-NLL. 

\subsection{Dataset-specific results for monolingual \texttt{GPT2-medium}}
\label{gpt-medium}
\begin{figure}[!th]
    \centering
    \includegraphics[width=0.95\linewidth]{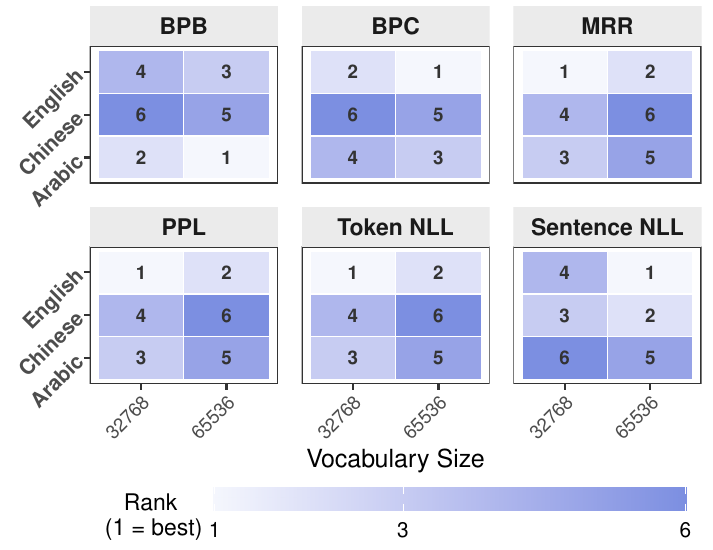}
   \caption{Language model rankings in a scaled setting (FLORES).}
    \label{fig:metrics_scale_flores}
\end{figure}

\begin{figure}[!th]
    \centering
    \includegraphics[width=0.95\linewidth]{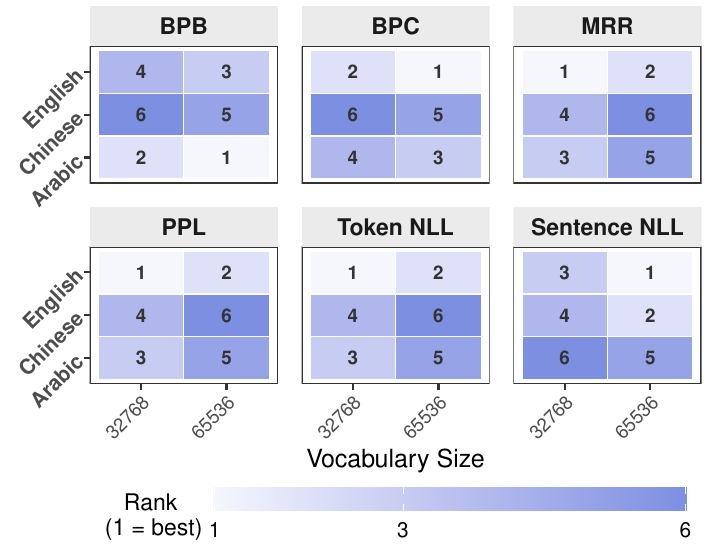}
   \caption{Language model rankings in a scaled setting (PUD).}
    \label{fig:metrics_scale_pud}
\end{figure}

\begin{figure}[!th]
    \centering
    \includegraphics[width=0.95\linewidth]{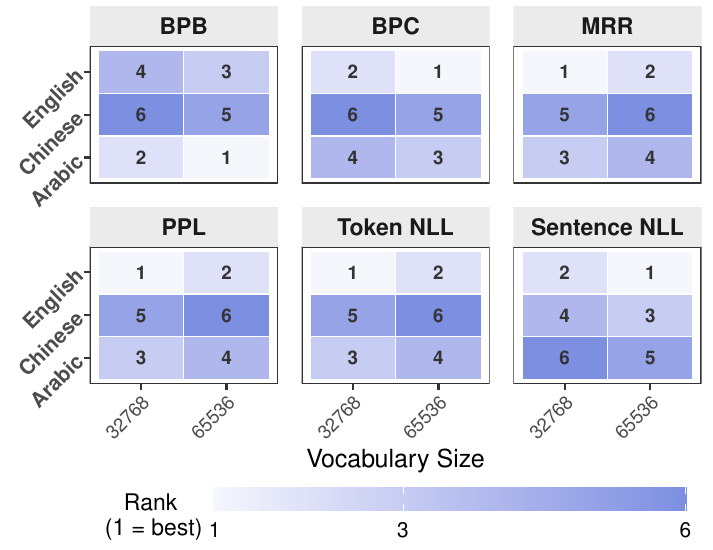}
   \caption{Language model rankings in a scaled setting (Parallel-3).}
    \label{fig:metrics_scale_test}
\end{figure}

Scaling experiment results on different datasets can be found in \Cref{fig:metrics_scale_flores,fig:metrics_scale_pud,fig:metrics_scale_test}.

\section{Paraphrases encode different amounts of information}
\label{sec:paraphrase}
Another concern may arise from our exclusion of paraphrases in Section~\ref{exp2}. Including paraphrases from \citet{freitag-etal-2020-human} could indeed change the results. For instance, DE$_{lowest}$ might obtain lower values and potentially rank below English, based on the results from \citet{poelman2026form}. However, this doubt relies on a common, but flawed, assumption that paraphrases within the same language should receive similar probabilities from a language model because they express the \textit{same meaning}, i.e., they are assumed to be semantically identical \citep{barzilay-mckeown-2001-extracting,boonthum-2004-istart,madnani-dorr-2010-generating} and therefore language models should assign similar probabilities to paraphrases of the same text \citep{poelman2026form}.\footnote{For example, \citet{poelman2026form} assume that paraphrases have the same information content and therefore language models should assign ``roughly the same [Token NLL or BPC] values for paraphrases within a single language''.}


We argue that this assumption does not necessarily hold. Paraphrases are often semantically similar but not strictly identical, and they encode different amounts of information. Consequently, there is no theoretical reason to expect that language models should assign more similar probabilities to two paraphrases than to translations of either sentence.

To illustrate this point, we use two paraphrases of a sentence from \citet{poelman-etal-2025-confounding} as an example:

\ex. \label{paraphrase_sentences}
\a. Do you know how to play chess? 
\b. Can you play chess?  


``Know how to play chess'' entails possessing the procedural knowledge about the rules of chess. ``Can play chess'' entails knowledge of the rules, but it is more sensitive to context. It could refer to possession of a chessboard or whether it is socially acceptable to play.





While in many contexts these differences may not be important, paraphrases are rarely semantically equivalent. Even grammatical alternations, e.g. dative alternation (c.f. ``I threw the box to John'' and ``I threw John the box.''; \citealp[]{levin1993english}), may not be considered semantically equivalent  \citep{bresnan2009gradience}. 
It may be more precise to label the sentences in Ex. \ref{paraphrase_sentences} as \textit{quasi-paraphrases} \citep{bhagat-hovy-2013-squibs} or \textit{near-equivalence} \citep{santos2025paraphrase}, which may encompass a ``broader, approximate, equivalence''.

Admittedly, both translations and paraphrases may encode different amounts of information for the same source. However, paraphrases typically permit substantially greater variation because they are not constrained by crosslinguistic correspondence \citep[e.g.,][]{neusner1986translation}. As a result, inconsistencies observed with paraphrases do not necessarily undermine crosslingual evaluation. Instead, they introduce an additional and less controlled source of variation.

\section{Case study with \texttt{GPT2}}
\label{app:case_study}

To to illustrate the bias on perplexity introduced by differences in compression, we provide a case study using \texttt{openai/GPT-2} model (monolingual English) evaluated on four languages from FLORES. As shown in Table~\ref{tab:flores_results}, token-level metrics (PPL, token-NLL, MRR) all suggest that the model performs best on Russian. This is counterintuitive given that the model is trained only on English, and MultiBLiMP results (Table~\ref{tab:multiblimp_results})\footnote{For Chinese, it is evaluated on ZhoBLiMP \citep{liu2026systematic} \textsc{VerbPhrase} category.} show that GPT2 performs well only on English (0.97) and near chance on other languages (in fact, below chance in Russian). In contrast, Sentence-level NLL assigns the lowest score to English, aligning better with expectations and the MultiBLiMP evaluations.

\begin{table}[!th]
\small
\centering
\begin{tabular}{lc}
\toprule
Language & Accuracy \\
\toprule
English & 0.97 \\
Chinese & 0.54 \\
Russian & 0.43 \\
Arabic & 0.50 \\
\midrule
\end{tabular}
\caption{MultiBLiMP results of GPT-2 across four languages.}
\label{tab:multiblimp_results}
\end{table}

\begin{table}[!th]
\small
\centering
\begin{tabular}{lcccc}
\toprule
Language & PPL & Token-NLL & MRR & Sent-NLL \\
\toprule
English & 49.77 & 3.91 & 0.43 & \textbf{99.34} \\
Chinese & 32.83 & 3.49 & 0.36 & 277.67 \\
Russian & \textbf{9.45} & \textbf{2.25} & \textbf{0.51} & 248.56 \\
Arabic & 12.22 & 2.50 & 0.44 & 251.00 \\
\bottomrule
\end{tabular}
\caption{crosslingual evaluation results of an English language model on FLORES.}
\label{tab:flores_results}
\end{table}

\begin{table}[!th]
\footnotesize
\centering
\begin{tabular}{l|p{5cm}}
\toprule
Model & Training data \\
\midrule
Llama-3.1-8B & Primarily English; AR, RU, ZH not officially supported \citep{grattafiori2024llama} \\
\midrule
Llama-3.2-1B & Primarily English; AR, RU, ZH not officially supported \citep{grattafiori2024llama} \\
\midrule
Llama-3.2-3B & Primarily English; AR, RU, ZH not officially supported \citep{grattafiori2024llama} \\
\midrule
Qwen3-8B     & 36T tokens across 119 languages; primarily in English and Chinese \citep{yang2025qwen3} \\
\midrule
Teuken-7B     & Supporting 24 European languages; AR, RU, ZH not officially supported \citep{ali2024teuken} \\
\midrule
BLOOM-1.1B   & 45 languages; AR: <1\%, EN: 31.3\%, RU: 0\%, ZH: 18.3\% \citep{workshop2022bloom} \\
\midrule
BLOOM-7.1B   & 45 languages; AR: AR: <1\%, EN: 31.3\%, RU: 0\%, ZH: 18.3\%  \citep{workshop2022bloom} \\
\midrule 
XGLM-4.5B & 31 languages; training data EN>RU>ZH>AR \citep{lin-etal-2022-shot}\\
\midrule 
mGPT & 61 languages; training data EN>RU>ZH>AR \citep{shliazhko-etal-2024-mgpt} \\
\bottomrule
\end{tabular}
\caption{Models used in this study and their pretraining data composition.}
\label{tab:data_comp}
\end{table}

\begin{table*}[]
\small 
    \centering
    \begin{tabular}{c|cccc}
    \toprule 
     Languages & Language Family & Morphology & Orthography & Word Order \\
     \midrule 
     English  & Indo-European (Germanic) & Analytic        & Alphabetic (Latin)      & SVO \\
     
     French   & Indo-European (Romance)   & Fusional        & Alphabetic (Latin)      & SVO \\
     German   & Indo-European (Germanic)  & Fusional        & Alphabetic (Latin)      & SVO/V2 \\
     Polish   & Indo-European (Slavic)    & Fusional        & Alphabetic (Latin)      & Flexible (SVO) \\
     Finnish  & Uralic                    & Agglutinative   & Alphabetic (Latin)      & SVO \\
     Turkish  & Turkic                    & Agglutinative   & Alphabetic (Latin)      & SOV \\
     Chinese  & Sino-Tibetan              & Isolating       & Logographic             & SVO \\
     Arabic   & Afro-Asiatic (Semitic)    & Nonconcatenative& Abjad                   & VSO/SVO \\
     Korean   & Koreanic                  & Agglutinative   & Featural (Hangul)       & SOV \\
     \bottomrule
    \end{tabular} 
    \caption{Typological properties of the languages used in our experiments.}
    \label{typology}
\end{table*}
\section{Multilingual Experiment}
\label{app:multi}
\begin{figure*}
    \centering
    \includegraphics[width=\linewidth]{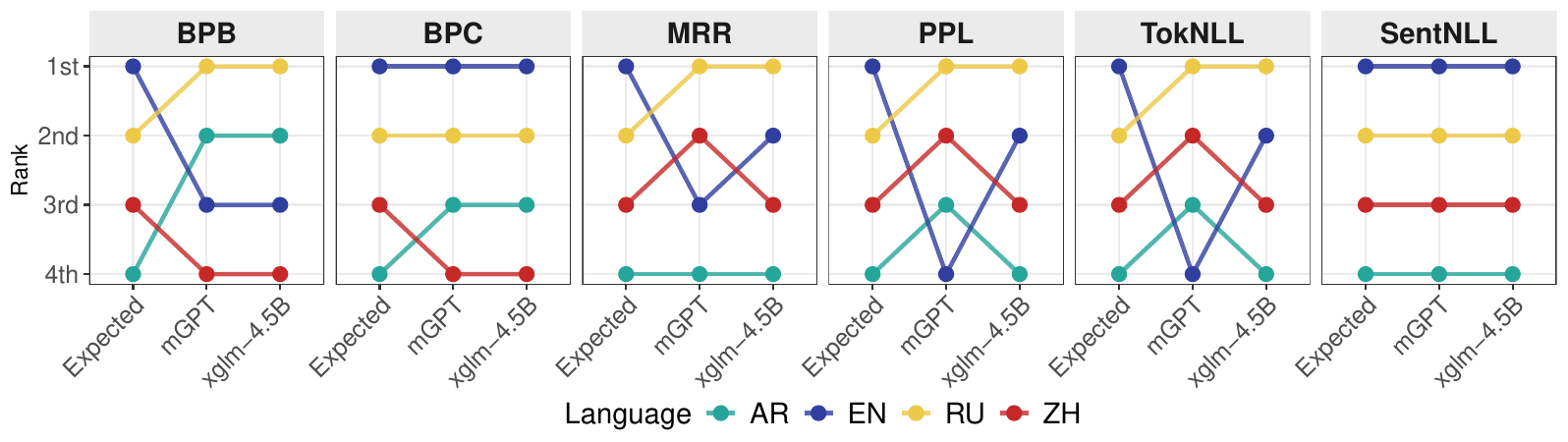}
    \caption{Expected ranking and actual ranking across 6 metrics for \texttt{mGPT} and \texttt{XGLM-4.5B}.}
    \label{fig:GLM}
\end{figure*}

\begin{figure*}
    \centering
    \includegraphics[width=\linewidth]{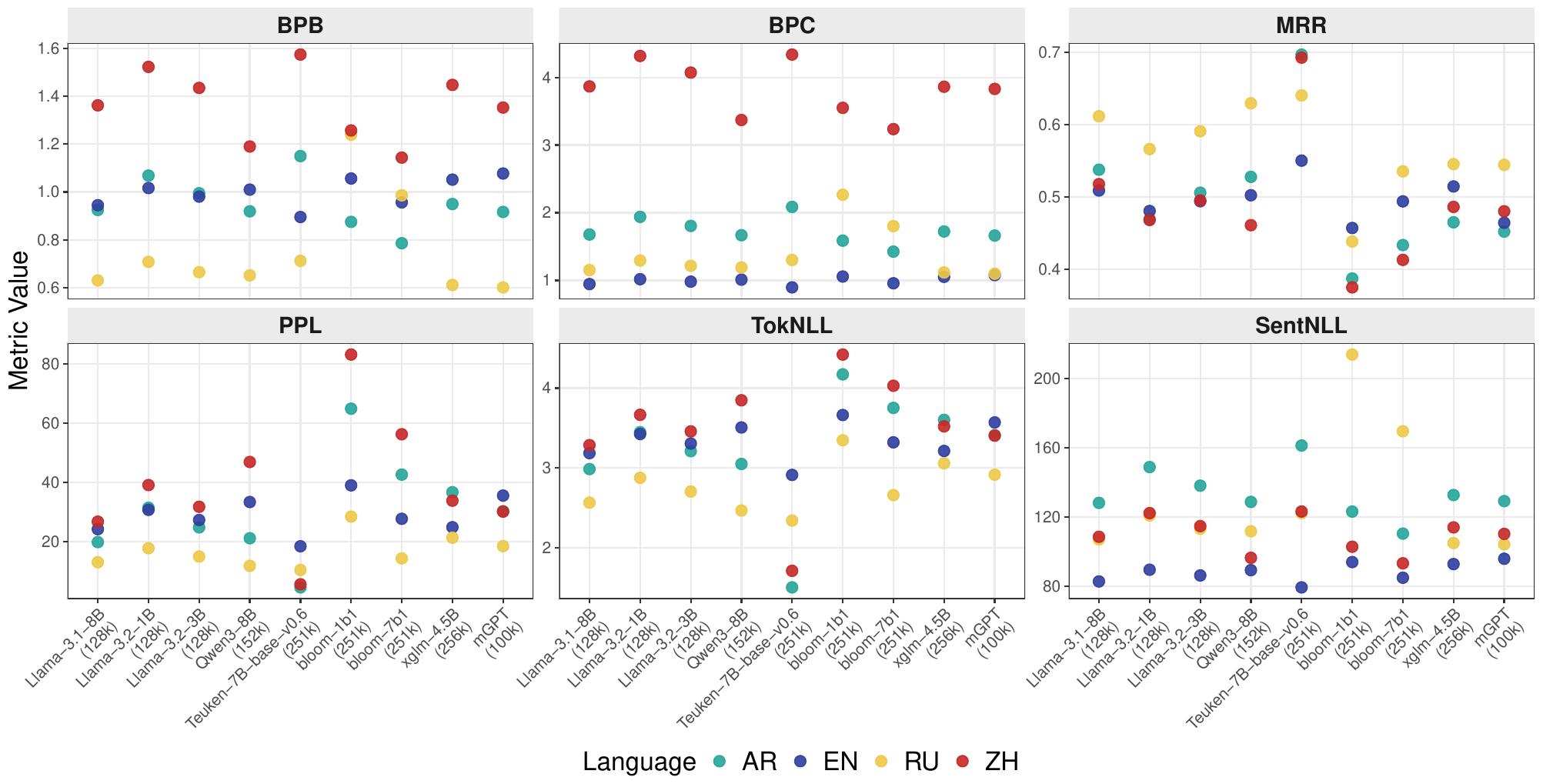}
    \caption{Raw results for all models across six evaluation metrics}
    \label{fig:raw_results}
\end{figure*}

Results for \texttt{mGPT} and \texttt{XGLM-4.5B} can be found in \Cref{fig:GLM} and the raw results for all multilingual evaluated can be found in \Cref{fig:raw_results}.

\end{document}